\documentclass[letterpaper, 10 pt, conference]{ieeeconf}  % Comment this line out if you need a4paper

\usepackage{kotex}
\IEEEoverridecommandlockouts                              % This command is only needed if 
\usepackage{amsmath, amssymb, mathtools} % assumes amsmath package installed
\usepackage{booktabs}  % \toprule, \midrule, \bottomrule
\usepackage{multirow}  % \multirow
\usepackage{textcase}
\usepackage{graphicx} 
\usepackage{balance}
\usepackage{bm}
\usepackage{placeins}

\newcommand{\best}[1]{\textbf{#1}}
\newcommand{\sbest}[1]{\underline{#1}}

\graphicspath{{figs/}}

\title{\LARGE \bf
Terrain-Enhanced Resolution-aware Refinement Attention for Off-Road Segmentation
}

\author{Seongkyu Choi$^{1}$ and Jhonghyun An$^{1}$% <-this % stops a space
\thanks{$^{1}$Gachon University, Seongnam, Republic of Korea
        {\tt\small seongkyu950324@gachon.ac.kr}}%
\thanks{$^{1}$Gachon University, Seongnam, Republic of Korea
        {\tt\small jhonghyun@gachon.ac.kr}}%
\thanks{Corresponding author: Jhonghyun An}%
}
\begin{document}

\maketitle
\thispagestyle{empty}
\pagestyle{empty}

%%%%%%%%%%%%%%%%%%%%%%%%%%%%%%%%%%%%%%%%%%%%%%%%%%%%%%%%%%%%%%%%%%%%%%%%%%%%%%%%
\begin{abstract}
Off-road semantic segmentation suffers from thick, inconsistent boundaries, sparse supervision for rare classes, and pervasive label noise. Designs that fuse only at low resolution blur edges and propagate local errors, whereas maintaining high-resolution pathways or repeating high-resolution fusions is costly and fragile to noise. We introduce a resolution-aware token decoder that balances global semantics, local consistency, and boundary fidelity under imperfect supervision. Most computation occurs at a low-resolution bottleneck; a gated cross-attention injects fine-scale detail, and only a sparse, uncertainty-selected set of pixels is refined. The components are co-designed and tightly integrated: global self-attention with lightweight dilated depthwise refinement restores local coherence; a gated cross-attention integrates fine-scale features from a standard high-resolution encoder stream without amplifying noise; and a class-aware point refinement corrects residual ambiguities with negligible overhead. During training, we add a boundary-band consistency regularizer that encourages coherent predictions in a thin neighborhood around annotated edges, with no inference-time cost. Overall, the results indicate competitive performance and improved stability across transitions.
\end{abstract}

% \begin{figure}[t]
%   \centering
%   \includegraphics[width=\linewidth]{figure/figure6.png}
%   \caption{Architectural comparison for off-road semantic segmentation.
% (a) \textbf{Standard}: a conventional encoder--decoder.
% (b) \textbf{Class-aware Attention}: class-conditioned token attention at the bottleneck.
% (c) \textbf{Proposed (ours)}: global--local token refinement, a single gated HR cross-attention, and sparse uncertainty-driven refinement. 
% All variants share the same backbone and decoder; only the bottleneck modules differ.}
%   \label{fig:terra_overview}
% \end{figure}

%%%%%%%%%%%%%%%%%%%%%%%%%%%%%%%%%%%%%%%%%%%%%%%%%%%%%%%%%%%%%%%%%%%%%%%%%%%%%%%%
\section{Introduction}

\begin{figure}[t]
  \centering
  \includegraphics[width=\linewidth]{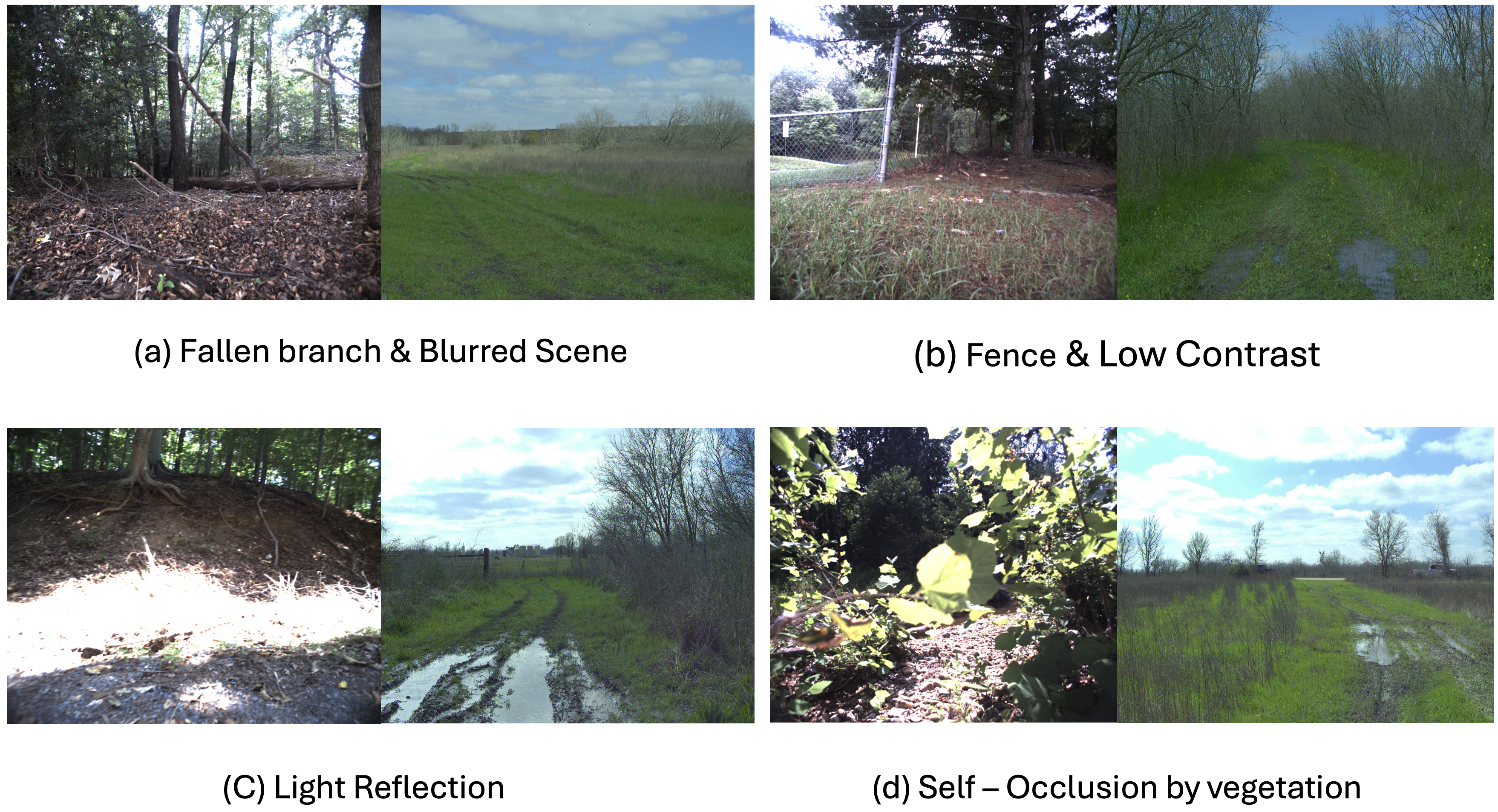}
  \caption{Off-road scenes pose four recurring challenges: (i) diffuse, platform-dependent transitions; (ii) frequent thin structures; (iii) low contrast under shadows, glare, and dust; (iv) vegetation self-occlusion and fine-scale variation.}
  \label{fig:terra_overview}
\end{figure}

Off-road semantic segmentation operates in highly irregular and heterogeneous scenes, where dense and precise real-world ground truth (GT) is inherently difficult to obtain~\cite{wigness2019rugd,jiang2021rellis}. Boundaries are thick and inconsistent, supervision for rare classes is sparse or absent, and temporal agreement across frames is weak. As illustrated in Fig.~\ref{fig:terra_overview}, (i) semantic transitions are diffuse and platform-dependent (grass vs.\ sparse vegetation, wet soil vs.\ shallow water)~\cite{wigness2019rugd,jiang2021rellis}, (ii) thin structures (stems, wires, fences) occur frequently~\cite{wigness2019rugd}, (iii) contrast is low due to shadows, glare, and dust under strong seasonal and illumination changes~\cite{wigness2019rugd,jiang2021rellis}, and (iv) vegetation exhibits self-occlusion and fine-scale variation~\cite{wigness2019rugd,jiang2021rellis}. These factors induce annotator disagreement and label drift within sequences, rendering pixel-accurate GT impractical in cost, time, and safety~\cite{wigness2019rugd,jiang2021rellis}. This is intrinsic to off-road perception rather than a defect of particular datasets such as RUGD~\cite{wigness2019rugd} and RELLIS-3D~\cite{jiang2021rellis}.

Under such supervision, standard encoder--decoder architectures and query-based decoders exhibit common limitations~\cite{wu2019fastfcn,xie2021segformer,zheng2021rethinking}.
(a) Early fusion at a low-resolution bottleneck, even with learned upsampling, weakens high-frequency cues needed for thin structures and sharp edges~\cite{wu2019fastfcn,wu2020cgnet,yan2021psp,peng2020semantic,huang2021recognizing}.
(b) When supervision emphasizes per-token \emph{classification evidence}, neighbor consistency and boundary--band interactions are under-constrained, leading to background bias and boundary-noise propagation; post-hoc regularizers like AAF and dense CRFs have aimed to mitigate this~\cite{ke2018adaptive,krahenbuhl2011efficient}.
(c) Auxiliary branches that are active only during training and removed at inference (e.g., the auxiliary losses in PSPNet/DeepLab families) introduce a train--test mismatch, destabilizing predictions in ambiguous regions~\cite{yan2021psp,peng2020semantic}.
In short, with imperfect and noisy labels under compute constraints, designs must preserve boundaries while jointly maintaining global semantics and local consistency.

\begin{figure*}[t]
  \centering
  \includegraphics[width=\textwidth]{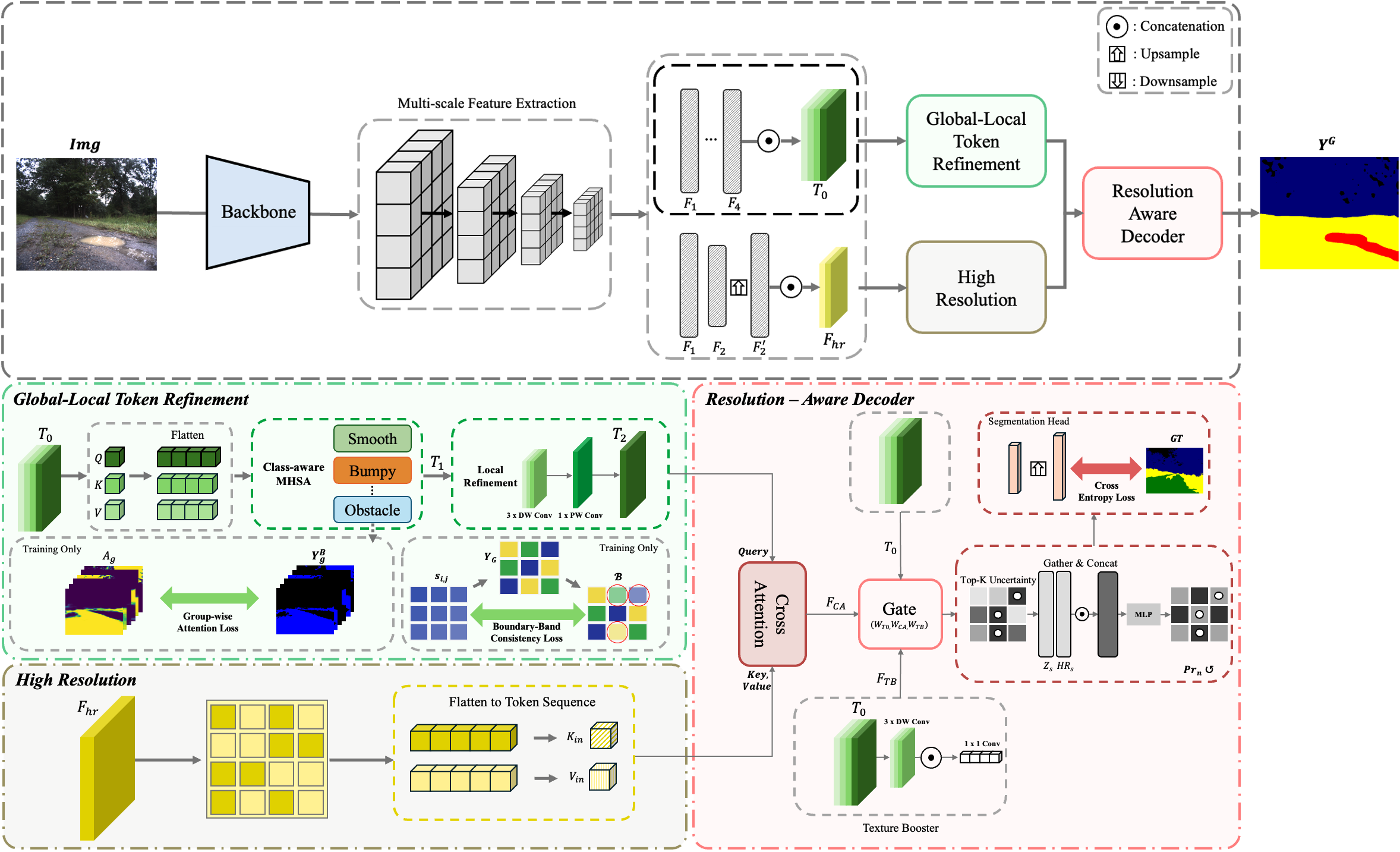}
  \caption{Overall framework. Multi-scale features are fused at a bottleneck and refined by GLTR. A single high-resolution (HR) cross-attention injects sharp cues, and a three-way gate blends $\{T_0, C, B\}$. CAPR revisits only uncertainty-selected pixels. During training, diagonal supervision and a thin boundary-band loss (BBL) regularize attention near edges.}
  \label{fig:overview}
\end{figure*}

We propose a \emph{resolution-aware token decoder} guided by two practical observations. 
First, global context is learned more stably on the low-resolution bottleneck lattice via attention~\cite{vaswani2017attention}, while local refinement is better handled by lightweight dilated depthwise convolutions that are less prone to propagating label noise~\cite{wu2019fastfcn}. 
Second, rather than repeatedly fusing at full resolution, consulting high-resolution cues once at the bottleneck and mixing them through a learnable gate preserves edge evidence with lower variance~\cite{sun2019hrnet}. 
In practice, this yields a simple recipe centered on our \emph{Resolution-Aware Decoder}, with full architectural details in Fig.~\ref{fig:overview}.

Our design comprises three co-designed, tightly integrated parts. 
(i) \emph{Global--Local Token Refinement (GLTR)} stabilizes bottleneck semantics via global self-attention followed by lightweight dilated depthwise refinement~\cite{vaswani2017attention,wu2019fastfcn}. 
(ii) The \emph{Resolution-Aware Decoder} concentrates computation at the bottleneck and consults a high-resolution feature via gated cross-attention~\cite{vaswani2017attention,sun2019hrnet}, while performing \emph{Class-Aware Point Refinement (CAPR)}, which sparsely re-evaluates uncertainty-selected pixels so that computational effort scales with uncertainty rather than image size~\cite{kirillov2020pointrend}. 
(iii) A training-only \emph{Boundary-Band Consistency Loss (BBL)} encourages agreement within a thin band around annotated edges, complementing evidence-centric supervision without adding inference cost. 
Together, these components balance global semantics, local consistency, and boundary fidelity under imperfect labels.

In experiments, the decoder attains competitive accuracy on RUGD and RELLIS-3D (6-class)~\cite{wigness2019rugd,jiang2021rellis}, with qualitatively more stable class transitions and fewer speckle artifacts in fine textures. 
Predictions remain aligned with RGB evidence even in regions affected by annotation artifacts, reflecting the effect of boundary-band regularization and the selective refinement within the Resolution-Aware Decoder.

The contributions of this paper are summarized as follows:
\begin{itemize}
  \item We propose a \emph{resolution-aware token decoder} that performs most computation on a low-resolution bottleneck and injects a single gated high-resolution cue; it integrates \emph{GLTR} global self-attention followed by lightweight dilated depthwise refinement to balance global semantics and local consistency~\cite{vaswani2017attention,wu2019fastfcn}.
  \item We introduce \emph{CAPR}, which sparsely re-evaluates only top-$K$ uncertain pixels so that decoding cost scales with uncertainty rather than image size~\cite{kirillov2020pointrend}.
  \item We add a training-only \emph{BBL} that supervises neighbor interactions in a thin band around annotated edges, improving transition stability with no inference-time overhead.
\end{itemize}

%%%%%%%%%%%%%%%%%%%%%%%%%%%%%%%%%%%%%%%%%%%%%%%%%%%%%%%%%%%%%%%%%%%%%%%%%%%%%%%%
\section{Related Work}

\subsection{Off-road Semantic Segmentation}
Camera-based off-road perception must parse unstructured terrain under imperfect labels and severe class imbalance (RUGD~\cite{wigness2019rugd}, RELLIS-3D~\cite{jiang2021rellis}).
Classic CNN decoders (PSPNet~\cite{yan2021psp}, DeepLabv3+~\cite{chen2018encoder}, DANet~\cite{fu2019dual}, OCRNet~\cite{huang2021recognizing}, PSANet~\cite{zhao2018psanet}) capture wide contextual information, but their early fusion at a low-resolution bottleneck blurs high-frequency details, weakening boundaries and thin structures.
Lightweight CNN variants BiSeNetV2~\cite{yu2021bisenet}, CGNet~\cite{wu2020cgnet}, FastSCNN~\cite{poudel2019fast}, FastFCN~\cite{wu2019fastfcn} achieve lower latency but remain fragile near noisy boundaries.
More recently, token and Transformer-based decoders SETR~\cite{zheng2021rethinking}, DPT~\cite{ranftl2021vision}, SegFormer~\cite{xie2021segformer}, SegNeXt~\cite{guo2022segnext} and class-aware designs (GA-Nav~\cite{guan2022ga}) have improved the trade-off between accuracy and efficiency.
However, in off-road settings, heavy or repeated high-resolution fusion often amplifies annotation noise and destabilizes training under thick, inconsistent boundaries~\cite{sun2019hrnet}. 
At the same time, reliance on a purely low-resolution token lattice under-constrains neighbor consistency in boundary bands, leading to leakage and background bias.
Finally, maintaining a persistent high-resolution branch further increases memory and latency, hindering embedded deployment on robotic platforms~\cite{sun2019hrnet}.
These limitations point to the need for a \emph{resolution-aware token decoder} that concentrates computation on the low-resolution bottleneck, injects a single high-resolution cue via gated cross-attention, and sparsely refines uncertainty-selected pixels to preserve boundaries while avoiding the overhead of full-resolution branches.

\subsection{Uncertainty-Guided Point Refinement}
Dense post-processing approaches such as Conditional Random Field (CRF)-style models, or affinity or contrastive regularizers, can refine predictions, but they are computationally heavy and often brittle under noisy labels. 
Recent sparse revisiting methods instead update only uncertain locations: PointRend~\cite{kirillov2020pointrend} adaptively samples low-confidence pixels for iterative refinement, and SegFix~\cite{yuan2020segfix} corrects boundary errors by consulting a separate boundary predictor.
While effective, these approaches are not explicitly class-aware and are typically applied only at inference, creating a mismatch between training and deployment.

To address this gap, we employ \emph{CAPR}, which selectively re-evaluates the top-$K$ least-confident pixels across stages~\cite{kirillov2020pointrend}. 
Each candidate is refined using both its current logits and local high-resolution features, ensuring class-consistent corrections even for rare or ambiguous regions. 
Because updates are restricted to a sparse uncertainty set, the added cost scales with uncertainty rather than image size, yielding negligible overhead. 
Applying \emph{CAPR} consistently during both training and inference further stabilizes behavior and recovers thin structures that would otherwise be lost.

\subsection{Boundary-Consistency Regularization}
Under imperfect supervision, many works augment per-pixel classification with neighborhood constraints to curb leakage around edges.
Affinity-based alignment (for example, adaptive affinity fields), boundary-aware or edge losses, and contrastive objectives impose local consistency in the prediction or feature space~\cite{yuan2020segfix}, yet their effectiveness can be sensitive to noisy labels and they often incur extra complexity when applied at full resolution.
Moreover, most regularizers supervise outputs rather than the interactions that generate them, leaving attention-space neighbor relations under-constrained in thin boundary bands~\cite{cheng2021boundary}.

We instead adopt a \emph{BBL} that targets a thin band around annotated edges and supervises neighbor interactions where leakage is most likely. 
By restricting regularization to boundary neighborhoods and computing it on the bottleneck lattice, \emph{BBL} complements evidence-centric supervision, reduces sensitivity to annotation noise, and adds no inference-time cost.
This training-only regularizer works in concert with the \emph{resolution-aware decoder} and \emph{CAPR} to improve transition stability without sacrificing efficiency.

%%%%%%%%%%%%%%%%%%%%%%%%%%%%%%%%%%%%%%%%%%%%%%%%%%%%%%%%%%%%%%%%%%%%%%%%%%%%%%%%
\begin{table}[t]
  \centering
  \caption{\protect\NoCaseChange{Grouping of fine-grained RUGD labels into a 6-class hierarchy by surface texture and Semantic Segmentation.}}
  \label{tab:terrain-groups}
  \renewcommand{\arraystretch}{1.05}
  \begin{tabular}{@{}p{.36\columnwidth} p{.56\columnwidth}@{}}
    \toprule
    \textbf{Terrain Group} & \textbf{Representative Region Types} \\
    \midrule
    Smooth Region    & Concrete, asphalt \\
    Rough Region     & Gravel, grass, dirt, sand \\
    Bumpy Region     & Rock, rock bed \\
    Forbidden Region & Water, bushes, tall vegetation \\
    Obstacles        & Trees, poles, logs, etc. \\
    Background       & Void, sky, signs \\
    \bottomrule
  \end{tabular}
  \vspace{-6pt}
\end{table}

\begin{table}[t]
  \centering
  \caption{\protect\NoCaseChange{Grouping of fine-grained RELLIS-3D labels into a 6-class hierarchy by navigability and surface characteristics.}}
  \label{tab:terrain-groups}
  \renewcommand{\arraystretch}{1.05}
  \begin{tabular}{@{}p{.30\columnwidth} p{.62\columnwidth}@{}}
    \toprule
    \textbf{Terrain Group} & \textbf{Mapped Fine-Grained Labels} \\
    \midrule
    Smooth Region   
      & Concrete, asphalt \\
    Rough Region     
      & Dirt, grass \\
    Bumpy Region     
      & Mud, rubble \\
    Forbidden Region           
      & Water, bush \\
    Obstacles                  
      & Tree, pole, vehicle, object, etc. \\
    Background                 
      & Void, sky \\
    \bottomrule
  \end{tabular}
  \vspace{-6pt}
\end{table}

\begin{table*}[!t]
  \centering
  \small
  \setlength{\tabcolsep}{5.2pt}
  \renewcommand{\arraystretch}{1.1}
  \caption{Comparison on \textsc{RUGD} and \textsc{RELLIS-3D}. We report per-group IoU (\%), mean IoU (mIoU$\uparrow$), and average accuracy (aAcc$\uparrow$). Asterisks (*) denote transformer-based methods. Best per dataset in \textbf{bold}, second best \underline{underlined}.}
  \label{tab:sota}
  \begin{tabular}{@{}cl*{8}{c}@{}}
    \toprule
    \textbf{Dataset} & \textbf{Methods (IoU)} &
    \textbf{Smooth} & \textbf{Rough} & \textbf{Bumpy} &
    \textbf{Forbidden} & \textbf{Obstacle} & \textbf{Background} &
    \textbf{mIoU$\uparrow$} & \textbf{aAcc$\uparrow$} \\
    \midrule
    \multirow[c]{15}{*}{\textbf{RUGD}}
    & PSPNet~\cite{yan2021psp}         & 48.62 & 88.92 & 69.45 & 29.07 & 87.98 & 78.29 & 67.06 & 92.85 \\
    & DeepLabv3+~\cite{chen2018encoder}     &  5.86 & 84.99 & 50.40 & 25.04 & 87.50 & \best{81.47} & 55.88 & 91.51 \\
    & DANet~\cite{fu2019dual}          &  2.26 & 81.47 &  8.69 & 15.00 & 82.54 & 74.86 & 44.14 & 88.81 \\
    & OCRNet~\cite{huang2021recognizing}         & 66.29 & 89.47 & 76.15 & 59.14 & 88.77 & 79.17 & 76.50 & 93.46 \\
    & PSANet~\cite{zhao2018psanet}         & 34.92 & 87.70 & 35.64 &  8.66 & 86.95 & 78.97 & 55.47 & 92.13 \\
    & BiSeNetv2~\cite{yu2021bisenet}      & 24.27 & 89.99 & \sbest{89.99} & 83.31 & 90.93 & 75.29 & 75.10 & 93.40 \\
    & CGNet~\cite{wu2020cgnet}          & 40.84 & 90.39 & 85.67 & 76.21 & 89.75 & 74.48 & 76.22 & 93.29 \\
    & FastSCNN~\cite{poudel2019fast}       & 83.03 & 92.82 & 87.69 & 81.05 & 90.94 & 75.11 & 85.11 & 94.77 \\
    & FastFCN~\cite{wu2019fastfcn}        & 26.27 & 89.85 & 85.95 & 84.13 & 91.23 & 75.63 & 75.51 & 93.46 \\
    & *SETR~\cite{zheng2021rethinking}          & 89.77 & 92.46 & 84.58 & 70.33 & 89.55 & 70.47 & 82.86 & 94.09 \\
    & *DPT~\cite{ranftl2021vision}           &  1.04 & 81.23 & 22.98 & 25.84 & 89.18 & 74.50 & 49.13 & 88.77 \\
    & *SegFormer~\cite{xie2021segformer}     & 93.26 & 93.16 & 87.56 & 77.31 & 91.20 & 78.50 & 86.83 & 95.17 \\
    & *SegNeXt~\cite{guo2022segnext}       & 90.39 & 91.17 & 83.96 & 65.43 & 87.80 & 68.17 & 81.15 & 93.22 \\
    & *GA-Nav~\cite{guan2022ga}        & \best{95.15} & \best{94.45} & 89.83 & \sbest{86.25} & \sbest{91.95} & 76.86 & \sbest{89.08} & \sbest{95.66} \\
    \midrule
    & \textbf{*TERRA (ours)}   & \sbest{94.56} & \sbest{94.21} & \best{90.19} & \best{86.40} & \best{92.37} & \sbest{79.90} & \best{89.60} & \best{95.85} \\
    \midrule
    \multirow[c]{15}{*}{\textbf{RELLIS-3D}}
    & PSPNet~\cite{yan2021psp}         & 69.21 & 80.99 &  8.89 & 53.70 & 60.70 & 94.67 & 61.36 & 86.01 \\
    & DeepLabv3+~\cite{chen2018encoder}     & 65.76 & 79.84 & 19.72 & 47.52 & 64.88 & 95.92 & 62.27 & 85.84 \\
    & DANet~\cite{fu2019dual}          & 72.93 & 85.18 & 13.10 & 60.60 & 70.53 & 95.65 & 66.38 & 89.11 \\
    & OCRNet~\cite{huang2021recognizing}         & 74.67 & 83.04 & 27.76 & 60.44 & 62.35 & 92.58 & 66.81 & 86.95 \\
    & PSANet~\cite{zhao2018psanet}         & 64.06 & 75.29 & 17.08 & 47.45 & 61.74 & 94.31 & 59.99 & 83.71 \\
    & BiSeNetv2~\cite{yu2021bisenet}      & 65.56 & 73.24 & 39.35 & 48.17 & 71.91 & 93.78 & 65.33 & 83.03 \\
    & CGNet~\cite{wu2020cgnet}          & 62.84 & 74.17 & 49.57 & 45.41 & 68.88 & 94.53 & 65.90 & 82.70 \\
    & FastSCNN~\cite{poudel2019fast}       & 67.06 & 77.60 & \best{56.49} & 49.76 & 70.31 & 94.43 & 69.27 & 84.51 \\
    & FastFCN~\cite{wu2019fastfcn}        & 70.51 & 79.15 & 49.72 & 51.37 & 63.90 & 94.82 & 68.24 & 84.10 \\
    & *SETR~\cite{zheng2021rethinking}          & 65.37 & 78.64 & 40.89 & 52.59 & 63.80 & 91.87 & 65.53 & 83.59 \\
    & *DPT~\cite{ranftl2021vision}           &  5.42 & 76.65 & 47.13 & 54.87 & 62.74 & 85.50 & 55.38 & 81.61 \\
    & *SegFormer~\cite{xie2021segformer}     & 60.28 & 79.78 & 53.35 & 53.78 & 70.15 & 94.37 & 68.62 & 85.37 \\
    & *SegNeXt~\cite{guo2022segnext}       & 51.67 & 78.40 & 19.38 & 42.61 & 66.04 & 92.05 & 58.36 & 82.16 \\
    & *GA-Nav~\cite{guan2022ga}        & \sbest{78.50} & \best{88.25} & \sbest{37.28} & \best{72.34} & \best{74.75} & \sbest{96.07} & \best{74.44} & \best{91.69} \\
    \cmidrule(lr){2-10}
    & \textbf{*TERRA (ours)}   & \best{80.68} & \sbest{87.12} & 31.96 & \sbest{70.63} & \sbest{74.64} & \best{96.11} & \sbest{73.52} & \sbest{91.18} \\
    \bottomrule
  \end{tabular}
\end{table*}

%%%%%%%%%%%%%%%%%%%%%%%%%%%%%%%%%%%%%%%%%%%%%%%%%%%%%%%%%%%%%%%%%%%%%%%%%%%%%%%%

\section{Proposed Method}
\label{sec:method}

We now formalize our \emph{resolution-aware decoder}: \emph{GLTR} on the bottleneck lattice, a single gated high-resolution cross-attention, \emph{CAPR} for sparse updates, and a training-only \emph{BBL}—as illustrated in Fig.~\ref{fig:overview}.

\subsection{Global--Local Token Refinement}

Off-road scenes often exhibit coarse textures and thickly annotated boundaries, so early full-resolution fusion tends to smear details. To obtain a stable representation, we first tokenize an RGB image $I \in \mathbb{R}^{3\times H\times W}$ into $7\times7$ patches with stride $4$ and feed them to the first transformer stage; each following stage performs $2\times2$ patch merging (stride $2$), halving the spatial resolution along both height and width. After the $i$-th stage, the spatial size becomes $H_i \times W_i = H/2^{\,i+1} \times W/2^{\,i+1}$, yielding four feature maps $f_1,\dots,f_4$ at strides $\{4,8,16,32\}$ with channels $\{32,64,160,256\}$.

To fuse features at the same spatial locality, we map each $f_i$ to a shared bottleneck resolution $(H_f, W_f)$ using a per-scale projection and bilinear resizing, then concatenate the aligned tensors along the channel axis and pass the result through a lightweight fusion layer to produce the initial token lattice $T_0$. Consolidating semantics at the bottleneck preserves multi-scale context while avoiding repeated high-resolution fusion and limiting noise amplification near thick and inconsistent boundaries.

We next capture long-range relations so that tokens absorb class-specific global cues. 
This is realized by multi-head self-attention (MHSA) applied on $T_0$. 
We first aggregate the per-head outputs and project them as
\begin{equation}
\label{eq:msa_heads}
Z \;=\;
\Big(
\operatorname*{Concat}_{h=1}^{H}
\big[
  \mathrm{softmax}\!\big(\tfrac{Q_h K_h^{\top}}{\sqrt{d_h}}\big)\, V_h
\big]
\Big) W_o
\end{equation}
and obtain the globally refined tokens via a residual connection:
\begin{equation}
\label{eq:msa_t1}
T_1 \;=\; T_0 \;+\; Z .
\end{equation}
Here $Q_h = T_0 W_q^{(h)}$, $K_h = T_0 W_k^{(h)}$, and $V_h = T_0 W_v^{(h)}$ are the query, key, and value projections for head $h$ with head dimension $d_h$; 
$\operatorname*{Concat}_{h=1}^{H}[\cdot]$ concatenates the $H$ head outputs along the channel axis, and $W_o$ is the output projection.

With global context established in \(T_1\), we restore local coherence using a lightweight refinement block with parallel dilated depthwise branches, followed by pointwise mixing and a nonlinearity, denoted by \(\phi(\cdot)\), resulting in \(T_2 = T_1 + \phi(T_1)\) that maintains global consistency while preserving thin structures and boundary continuity under noisy supervision.

\subsection{High-Resolution Cross-Attention with Gated Fusion}

Even after global/local refinement, thin structures and true boundaries can remain ambiguous without high-resolution cues. 
To preserve efficiency while avoiding repeated high-resolution fusion, we inject high-resolution information once via multi-head cross-attention (MHCA): the bottleneck tokens $T_2$ act as queries and the high-resolution feature $F_{\mathrm{HR}}$ provides keys/values, yielding an update $C$ that captures sharp, spatially precise evidence.

To avoid overusing high-resolution cues and to remain robust in homogeneous areas, we fuse three sources with a learnable gate: the stable bottleneck tokens $T_0$, the high-resolution update $C$, and a mid-frequency texture branch $B$. 
We compute a global summary vector from the bottleneck path only, $\bar T_0 \in \mathbb{R}^{C}$, and use it as the gating descriptor $\mathbf{z}=\bar T_0$. 
We then produce a three-way softmax gate over the sources $\{T_0,\ C,\ B\}$ by first computing per-branch energies 
$e_k=\langle \mathbf{u}_k,\mathbf{z}\rangle+\beta_k$ for $k\in\{1,2,3\}$, 
where $\mathbf{u}_k \in \mathbb{R}^{C}$ and $\beta_k \in \mathbb{R}$ are learnable parameters, with $k=1,2,3$ indexing the $\{T_0,\,C,\,B\}$ branches. 
The normalized gate weights are
\begin{equation}
\label{eq:gate_energy}
w_k \;=\; \frac{\exp(e_k)}{\sum_{j=1}^{3}\exp(e_j)}, \qquad k\in\{1,2,3\},
\end{equation}
which satisfy $w_k \ge 0$ and $\sum_{k=1}^{3} w_k = 1$. 
The final tokens are then
\begin{equation}
\label{eq:gate_mix}
\hat T \;=\; w_{T_0}\,T_0 \;+\; w_{C}\,C \;+\; w_{B}\,B,
\end{equation}
so that sharp boundaries or thin structures naturally upweight $C$ or $B$, while homogeneous regions favor $T_0$. 
Here $B$ denotes a shallow convolutional texture booster that restores mid-frequency detail.

\subsection{Class-Aware Point Refinement}

Because off-road data have thick boundaries and rare classes, refining all pixels at full resolution is wasteful and may propagate noise. 
CAPR balances accuracy and stability by refining only where the model hesitates. 
We measure uncertainty from upsampled logits using the top-2 probability margin 
\(m(i)=p_{[1]}(i)-p_{[2]}(i)\) with \(p_i=\mathrm{softmax}(Z_i^{\uparrow})\). 
We then select the set of most uncertain pixels as
\begin{equation}
\label{eq:capr_sel}
S \;=\; \mathrm{TopK}_{\,i}\!\big(-m(i),\,K\big),
\end{equation}
where $\mathrm{TopK}_{\,i}(-m(i),K)$ denotes the $K$ pixels with the smallest margins $m(i)$, 
i.e., the $K$ most uncertain locations according to the top-2 probability gap. 
For $i\!\in\!S$, we apply a small \emph{multi-layer perceptron (MLP)} over local HR features 
concatenated with current logits to produce a \emph{residual} correction. 
This concentrates computation near boundaries, fine structures, and rare classes, 
while leaving confident regions untouched, as shown in the bottom-right of Fig.~\ref{fig:overview}.

\subsection{Loss Functions}

Given thick boundaries and frequent label noise, our training objective jointly targets \emph{stable global semantics}, class–token alignment, and local consistency within a boundary band. We use three terms: (i) a standard cross-entropy segmentation loss $L_{\mathrm{seg}}$ to anchor global semantics at the logit level; (ii) a diagonal supervision term $L_{\mathrm{diag}}$ that aligns the class-aware attention’s diagonal channel with the ground-truth class at each location; and (iii) a boundary-band consistency term $L_{\mathrm{bbl}}$ that encourages locally consistent interactions only near annotated boundaries.

For the boundary-band consistency, let $\mathcal{B}$ denote the boundary band and $\mathcal{R}(i)$ the ring set adjacent to location $i$. Define the set of boundary–adjacent pairs as
\[
E=\{(i,j)\mid i\in\mathcal{B},\; j\in\mathcal{R}(i)\}.
\]
For each pair $(i,j)$, let $s_{ij}\in\mathbb{R}$ be the model’s \emph{same-class} logit score and $t_{ij}\in\{0,1\}$ the target (1 if $y_i{=}y_j$, else 0). Then
\begin{equation}
\label{eq:bbl}
L_{\mathrm{bbl}}
=\frac{1}{|E|}\sum_{(i,j)\in E}
\ell_{\text{bce}}\!\big(s_{ij},\,t_{ij}\big),
\end{equation}
where $\ell_{\text{bce}}$ denotes binary cross-entropy with logits. To avoid early over-regularization, we warm up the weight on $L_{\mathrm{bbl}}$ from a small value to a larger one, thereby increasing coupling within the boundary band as training confidence grows.

The full objective is
\begin{equation}
\label{eq:total}
\mathcal{L}
\;=\;
L_{\mathrm{seg}}
\;+\;
\lambda_{\mathrm{diag}}\,L_{\mathrm{diag}}
\;+\;
\lambda_{\mathrm{bbl}}(t)\,L_{\mathrm{bbl}}.
\end{equation}
This combination balances global and local cues, suppresses boundary noise, and strengthens class-aligned attention—key to robust off-road segmentation.

%%%%%%%%%%%%%%%%%%%%%%%%%%%%%%%%%%%%%%%%%%%%%%%%%%%%%%%%%%%%%%%%%%%%%%%%%%%%%%%%
\begin{figure}[t]
  \centering
  \includegraphics[width=\linewidth]{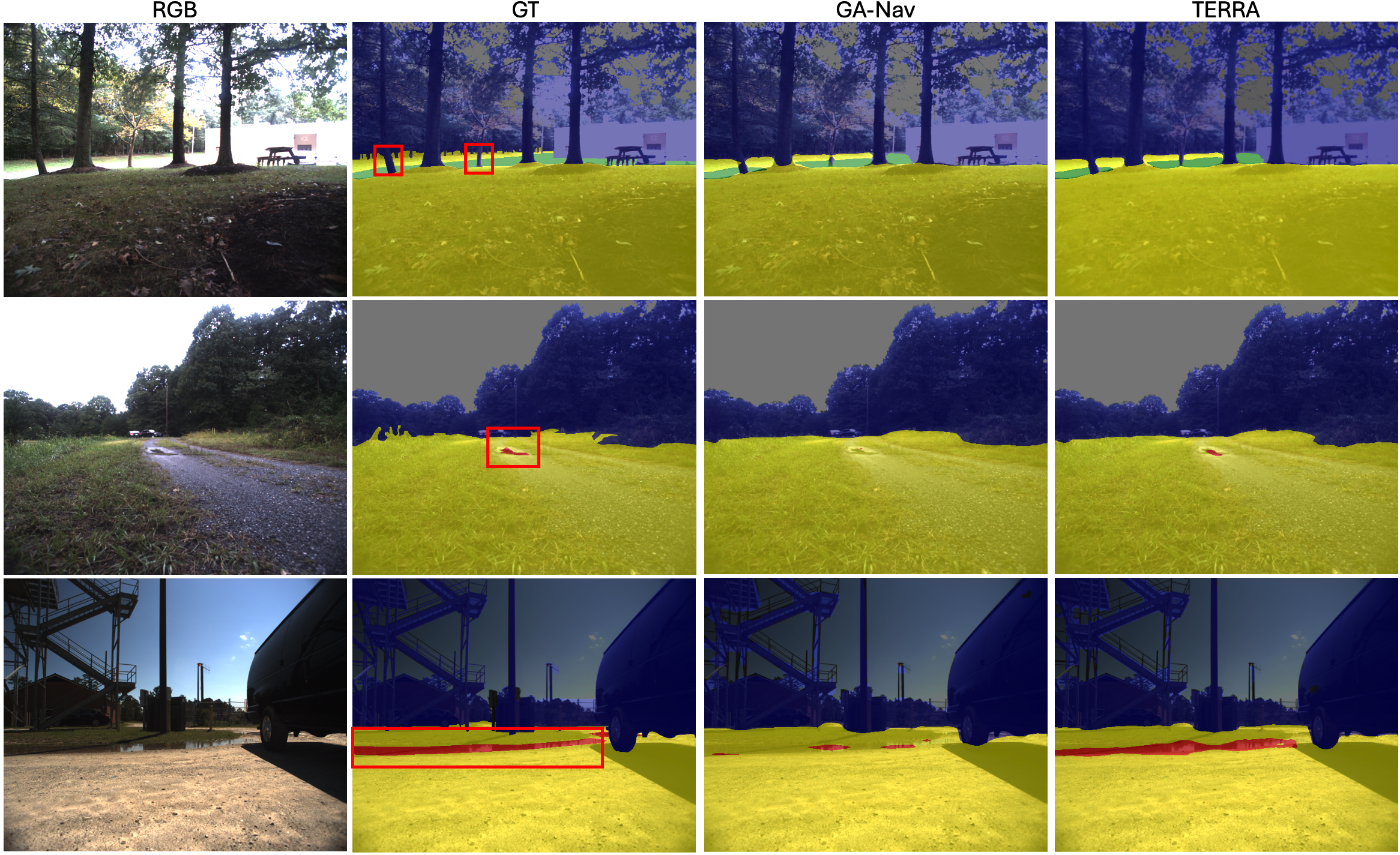}
  \caption{Qualitative comparison on RUGD with the baseline (left) and TERRA (right).
Red boxes highlight regions where the baseline either misses classes or mixes them. In contrast, TERRA captures thin structures and fine details more precisely, reduces interior holes and clutter, and traces boundaries more sharply and continuously—resulting in segmentations that better reflect the actual scene layout.}
  \label{fig:qual_rugd}
\end{figure}

\begin{figure}[t]
  \centering
  \includegraphics[width=\linewidth]{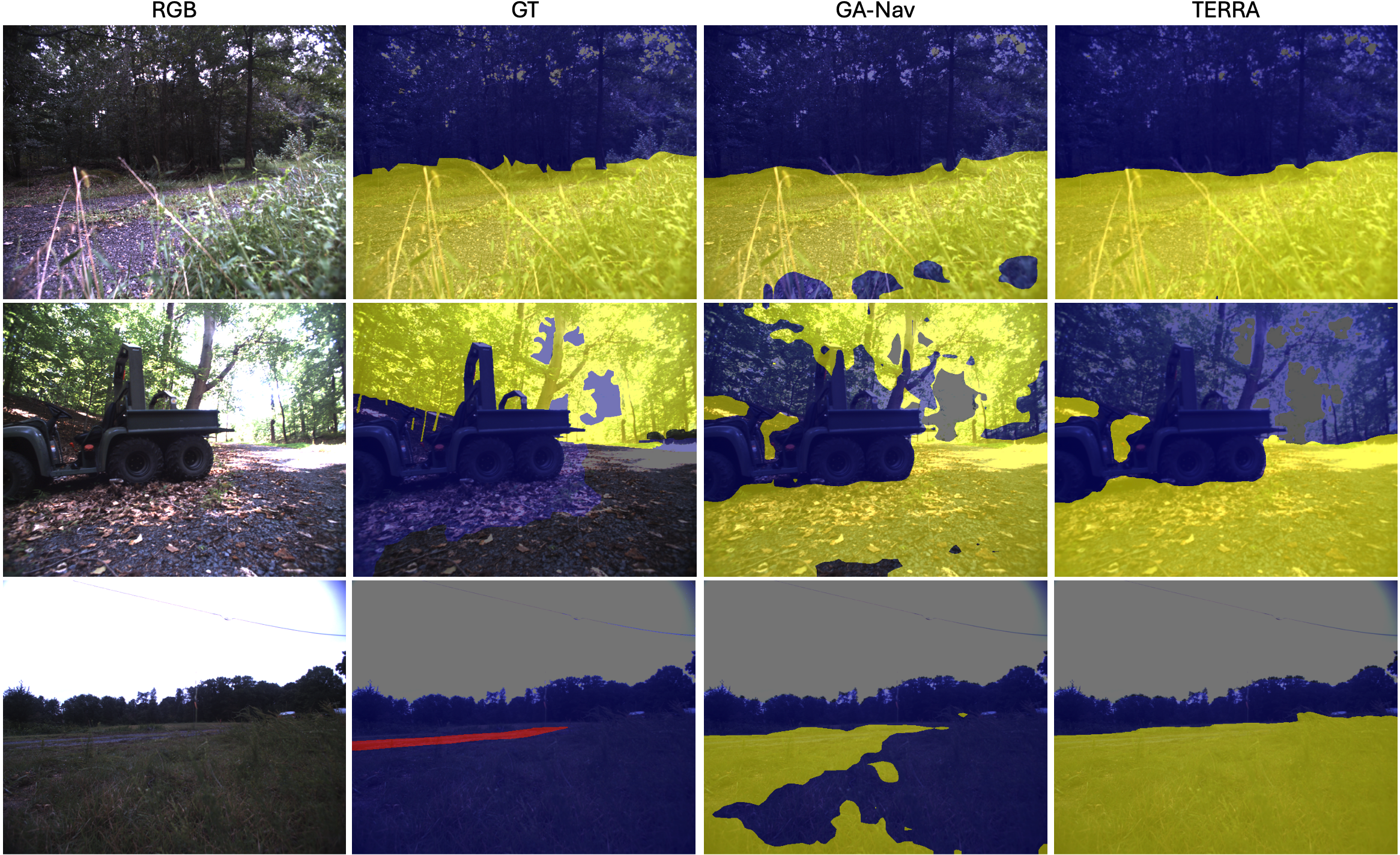}
  \caption{Qualitative comparison on RUGD with noisy ground-truth (GT).
GT labels misalign with actual scene structure, causing GA-Nav to inherit
labeling errors and produce fragmented regions. In contrast, TERRA learns
more robust representations, yielding cleaner borders and predictions that
better align with the true layout despite annotation noise.}
  \label{fig:qual_gtnoise}
\end{figure}

\section{Experiments}
\subsection{Datasets}
\emph{RUGD} offers RGB off-road scenes with diverse terrains (soil, grass, gravel, rocks, water, vegetation, man-made), thin structures, and irregular boundaries. We follow the official split and use RGB only. Class imbalance and boundary ambiguity motivate our boundary-band consistency regularizer and CAPR.

\emph{RELLIS-3D} is collected in Clearpath Warthog unmanned ground vehicle (UGV) environments. We use RGB-only. Four sequences (2k+ frames each): seq. 0--3 (11{,}497 frames) for training; seq. 4 for testing, divided into three routes (500/500/700 frames).

Both datasets are remapped to the GA-Nav 6-class mapping for all quantitative results.

\subsection{Metrics and Protocol}
We report mean Intersection-over-Union (mIoU), average accuracy (aAcc), and \emph{boundary IoU (bIoU)}.
The bIoU is computed only on a thin band around ground-truth boundaries (obtained from the GT boundary mask with a small morphological band), so it reflects boundary quality independently of interior regions~\cite{cheng2021boundary}.
All experiments follow the common 6-class mapping; the \textit{Background} group includes \emph{void}/\emph{sky}/\emph{signs} (excluded from loss, included in evaluation via mapping).
Inference is single-scale without test-time augmentation (no flip, no multi-scale): RUGD at $300{\times}375$, RELLIS-3D at $375{\times}600$.
Unless stated otherwise, methods use RGB-only input and identical preprocessing, cropping, and schedules across datasets.

\subsection{Main Results}
Table~\ref{tab:sota} summarizes cross-method comparisons on the GA-Nav 6-class mapping. On RUGD, our method attains the best overall scores with 89.60 mIoU and 95.85 aAcc, surpassing GA-Nav (89.08/95.66) and all CNN/Transformer baselines. Per-class, our method improves \emph{Background} by +3.04 IoU (76.86$\rightarrow$79.90), and also raises \emph{Obstacle} (+0.42) and \emph{Forbidden} (+0.15), while being slightly lower on texture-dominated classes \emph{Smooth}/\emph{Rough}/\emph{Bumpy} ($-0.59$/$-0.24$/$-0.64$). Qualitatively, as shown in Figs.~\ref{fig:qual_rugd} and \ref{fig:qual_gtnoise}, predictions exhibit fewer spurious fragments and reduced leakage into large traversable regions, yielding cleaner, more contiguous boundaries under label noise; Fig.~\ref{fig:qual_rugd} illustrates standard scenes, while Fig.~\ref{fig:qual_gtnoise} highlights robustness under noisy GT. On RELLIS-3D, our method reaches 73.52 mIoU and 91.18 aAcc, comparable to GA-Nav (74.44/91.69). Class-wise, our method is on par for \emph{Rough} (87.12 vs.\ 87.28) and slightly higher for \emph{Background} (96.11 vs.\ 96.07), but trails on \emph{Smooth}/\emph{Bumpy}/\emph{Forbidden}/\emph{Obstacle}, which lowers the average. Nonetheless, the qualitative results reveal fewer holes in large flat regions and sharper object contours, consistent with the single HR injection and the selective refinement of \emph{CAPR}; these trends are visually consistent on RELLIS-3D as well, as shown in Fig.~\ref{fig:qual_rugd}.

\subsection{Ablation Studies}

We conduct ablation experiments on RUGD to analyze the contribution of each component in our framework (Table~\ref{tab:ablation}). 
The baseline model achieves 88.32 mIoU and 40.07 bIoU, serving as a strong reference but leaving boundary quality relatively low.

Adding \emph{GLTR} stabilizes global semantics and slightly improves mIoU to 88.47 \,(+0.15), while boundary IoU shows minor fluctuation (39.90).
Introducing the \emph{Resolution-Aware Decoder} yields a clearer gain in boundary quality, with bIoU increasing to 40.55 \,(+0.65 from GLTR) and mIoU to 88.66 \,(+0.19).
The addition of \emph{CAPR} provides the largest incremental boost, raising performance to 89.58 mIoU \,(+0.92) and 43.97 bIoU \,(+3.42), indicating that selectively refining top-$K$ uncertain pixels is effective for rare classes and thin boundaries.
Finally, incorporating the \emph{BBL} during training offers additional regularization and smoother convergence, yielding 89.60 mIoU and 43.79 bIoU—comparable to \emph{CAPR} at the aggregate level.

Overall, these results show stepwise, complementary improvements rather than dramatic jumps: 
\emph{GLTR} consolidates global context, the \emph{resolution-aware decoder} injects HR detail once to enhance boundaries, \emph{CAPR} sparsely corrects uncertain predictions with negligible overhead, and \emph{BBL} regularizes boundary neighborhoods during training. 
We note that bIoU is reported only in the ablation to isolate boundary effects; cross-method main tables use mIoU/aAcc for fair comparison with prior work. 
For completeness, we include absolute scores alongside deltas and emphasize that the observed margins are modest; multi-seed runs and confidence intervals would further clarify statistical significance.

\begin{figure}[t]
  \centering
  \includegraphics[width=\linewidth]{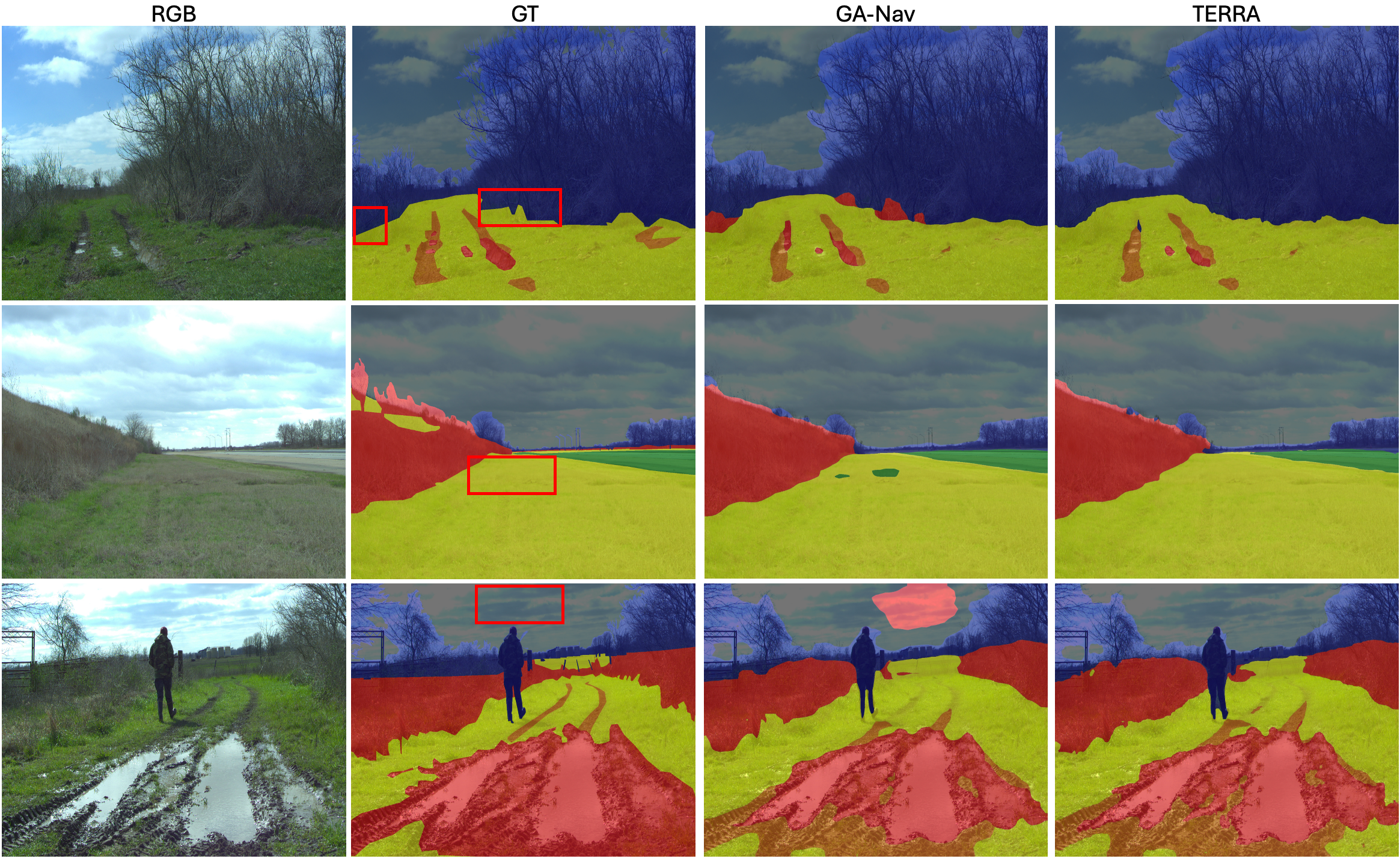}
  \caption{Qualitative comparison on \textsc{RELLIS-3D}. Columns: RGB, GT, GA-Nav, and \textbf{our method}. Compared with GA-Nav, our method suppresses small holes in wide traversable areas, reduces vegetation clutter, and yields sharper, more continuous boundaries despite annotation noise.}
  \label{fig:qual_rellis}
\end{figure}

\begin{table}[t]
\centering
\caption{Ablation on RUGD with incremental components.}
\label{tab:ablation}
\setlength{\tabcolsep}{6pt}
\begin{tabular}{lccc}
\hline
Variant & mIoU$\uparrow$ & bIoU$\uparrow$ & aAcc$\uparrow$ \\
\hline
Baseline                             & 88.32 & 40.07    & 95.44    \\
\,+ GLTR                             & 88.47 & 39.9     & 95.44 \\
\,+ Resolution-Aware Decoder         & 88.66 & 40.55    & 95.52    \\
\,+ CAPR \                           & 89.58 & 43.97    & 98.88    \\
\,+ BBL (training-only) \            & 89.60 & 43.79    & 98.88    \\
\hline
\end{tabular}
\end{table}

% \begin{table}[t]
% \centering
% \caption{Ablation of gate variants on RUGD.
% TB = TextureBooster, CA = HR cross-attention, T0 = bottleneck identity.}
% \label{tab:ablation_gate}
% \setlength{\tabcolsep}{8pt}
% \begin{tabular}{llcc}
% \toprule
% Group    & Combination         & mIoU$\uparrow$ & aAcc$\uparrow$ \\
% \midrule
% \multirow{3}{*}{Two-way}
%          & T0 + CA             & 89.76 & 95.88 \\
%          & TB + CA             & 89.78    & 95.84    \\
%          & TB + T0             & 89.8    & 95.91    \\
% \midrule
% Three-way& T0 + CA + TB        & 89.65 & 95.84 \\
% \bottomrule
% \end{tabular}
% \end{table}

% \usepackage{amssymb}  % for \checkmark, if not already included

% (Optional) Qualitative figures:
% \begin{figure*}[t]
%   \centering
%   \includegraphics[width=\textwidth]{figure/qual_rugd.png}
%   \caption{Qualitative comparison on RUGD. Red boxes highlight hole suppression and cleaner boundaries.}
%   \label{fig:qual1}
% \end{figure*}
% \begin{figure*}[t]
%   \centering
%   \includegraphics[width=\textwidth]{figure/qual_rellis.png}
%   \caption{Qualitative comparison on RELLIS-3D. Fewer spurious patches and more contiguous object boundaries.}
%   \label{fig:qual2}
% \end{figure*}

\section{CONCLUSIONS}

This paper presented \textsc{TERRA}, a resolution-aware token decoder designed for off-road semantic segmentation under noisy labels. By fusing multi-scale features in a stable bottleneck, injecting HR cues once with a three-way gate, and refining only uncertain pixels through CAPR, TERRA achieves a balance of global context, local detail, and boundary fidelity. A boundary-band loss further enhances robustness to annotation noise. Experiments on RUGD and RELLIS-3D confirm competitive or superior results over GA-Nav, showing cleaner boundaries and fewer artifacts, with potential to extend to other domains with coarse or unreliable annotations.

\bibliographystyle{ieeetr}
\bibliography{ref}

% \end{thebibliography}

\end{document}